\documentclass{article}

\usepackage[nonatbib, preprint]{neurips_2023}

\usepackage{hyperref}
\usepackage{url}
\usepackage{amsthm}
\usepackage{amsmath}
\usepackage[pdftex]{graphicx}
\usepackage{wrapfig}
\usepackage{algorithm}
\usepackage{algpseudocode} 

\theoremstyle{definition}

\usepackage[utf8]{inputenc} 
\usepackage[T1]{fontenc}    
\usepackage{hyperref}       
\usepackage{url}            
\usepackage{booktabs}       
\usepackage{amsfonts}       
\usepackage{nicefrac}       
\usepackage{microtype}      
\usepackage{xcolor}         

\title{Task Addition in Multi-Task Learning \\ by Geometrical Alignment}

\author{%
  Soorin Yim\\
  LG AI Research\\
  \texttt{soorin.yim@lgresearch.ai}\\
  \And
  Dae-Woong Jeong\\
  LG AI Research\\
  \texttt{dw.jeong@lgresearch.ai}\\
  \And
  Sung Moon Ko\\
  LG AI Research\\
  \texttt{sungmoon.ko@lgresearch.ai}\\
  \And
  Sumin Lee\\
  LG AI Research\\
  \texttt{sumin.lee@lgresearch.ai}\\
  \And
  Hyunseung Kim\\
  LG AI Research\\
  \texttt{hyunseung.kim@lgresearch.ai}\\
  \And
  Chanhui Lee\\
  LG AI Research\\
  \texttt{chanhui-lee@lgresearch.ai}\\
  \And
  Sehui Han\\
  LG AI Research\\
  \texttt{hansse.han@lgresearch.ai}\\
}

\newcommand{\gate}{{\it GATE}}

\begin{document}

\maketitle

\begin{abstract}
Training deep learning models on limited data while maintaining generalization is one of the fundamental challenges in molecular property prediction. 
One effective solution is transferring knowledge extracted from abundant datasets to those with scarce data.
Recently, a novel algorithm called Geometrically Aligned Transfer Encoder ({\gate}) has been introduced, which uses soft parameter sharing by aligning the geometrical shapes of task-specific latent spaces. 
However, {\gate} faces limitations in scaling to multiple tasks due to computational costs. 
In this study, we propose a task addition approach for {\gate} to improve performance on target tasks with limited data while minimizing computational complexity. 
It is achieved through supervised multi-task pre-training on a large dataset, followed by the addition and training of task-specific modules for each target task. 
Our experiments demonstrate the superior performance of the task addition strategy for {\gate} over conventional multi-task methods, with comparable computational costs.
\end{abstract}


\section{Introduction}
Molecular property prediction is a key area of computational chemistry, aimed at developing models that map molecular structures to their properties \cite{Deng2023}. These properties can vary from fundamental characteristics such as electron affinity and critical compressibility factor to complex and specific properties like chromophore lifetime. These predictions are crucial for accelerating the development of next-generation materials, optimizing chemical synthesis, and understanding molecular interactions.

However, accurately predicting molecular properties poses a significant challenge due to the complex relationships between molecular structures and their properties. Transfer learning has emerged as an essential technique to address the scarcity of labeled data and the high dimensionality of feature spaces in this field \cite{SHEN201929}.

By leveraging knowledge gained from related tasks or domains, transfer learning enables models to generalize better to target tasks with limited training data, enhancing prediction accuracy and robustness. This is particularly beneficial in molecular property prediction, where experimental data collection is often expensive and time-consuming. 

In the context of molecular property prediction, Geometrically Aligned Transfer Encoder ({\gate}) has been introduced as a promising foundation approach for knowledge transfer \cite{ko2024geometrically, ko2024multitask}. {\gate} aligns the geometrical shapes of latent spaces across tasks to transfer mutual information. However, {\gate}'s pairwise transfer between tasks results in $O(N^2)$ computational complexity as the number of tasks increases, making it computationally expensive. 

To address this issue, we propose a task addition approach for {\gate}, a two-stage framework that reduces the computational cost while maintaining effective knowledge transfer. In the first stage, supervised pre-training is performed on a large dataset. In the second stage, modules for each target task are added and trained on smaller datasets of the target task while the parameters pre-trained on source tasks remain fixed. This approach allows pre-trained models to be reused for diverse target tasks, significantly reducing training time.

In this paper, we introduce task addition for {\gate} to minimize computational cost while maintaining generalizability. The key contributions of this work are as follows:
\begin{itemize}
    \item We extend {\gate} with a task addition approach.
    \item Task-added {\gate} outperforms single-task learning (SINGLE) and task-added multi-task learning (MTL) across various molecular property prediction tasks.
    \item The training time for task-added {\gate} is significantly faster than training MTL models from scratch and comparable to SINGLE and task-added MTL.
    \item Our results show that task-added {\gate} is less dependent on the choice of source tasks, unlike task-added MTL, which heavily depends on source tasks.
\end{itemize}


\section{Task Addition by Geometrical Alignment}
Geometrical alignment is an effective method in a multi-task prediction setups. However, as shown in \cite{ko2024multitask}, the algorithm requires significant computational power as the number of tasks increases. To address this, we hereby introduce a specific mathematical description of the task addition method for {\gate} to accelerate the algorithm with minimal loss of prediction power.

In the {\gate} algorithm, the core assumption is that the essence of prediction performance lies in the geometrical characteristics of the corresponding latent space of a given input. In multi-task learning, different downstream tasks induce various latent spaces, yet the input remains equivalent for molecular property prediction tasks. Therefore, the primary strategy is to align the geometrical shapes of latent spaces from different prediction tasks to maximize the utilization of mutual information.

\begin{figure*}[!bt]
\begin{center}
\includegraphics[width=0.92\linewidth]{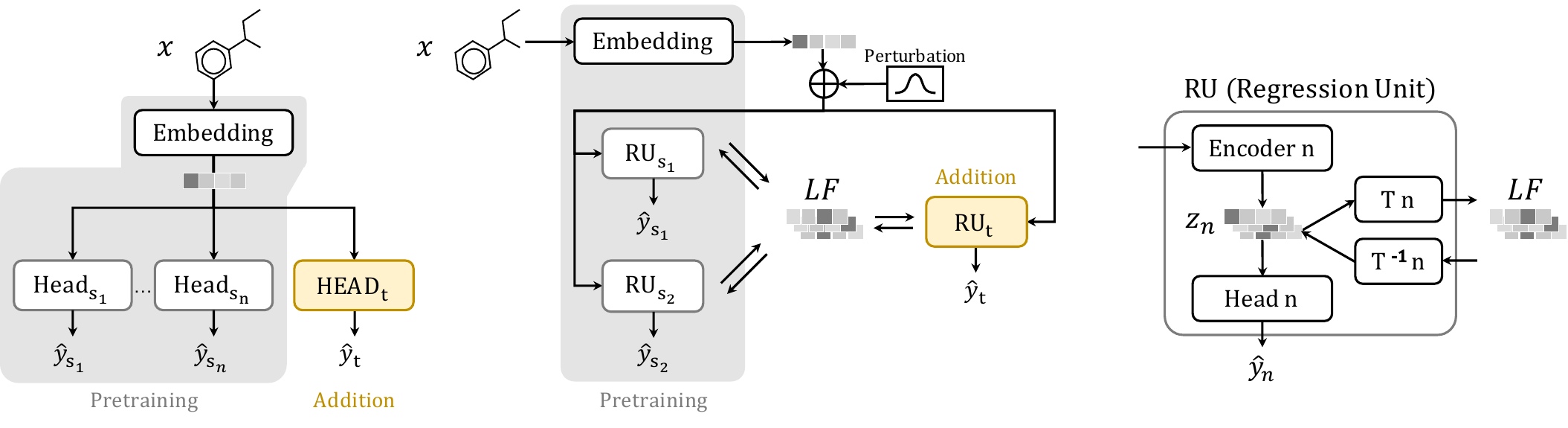}
\end{center}
\caption{Schematic diagram of task addition in multi-task learning, where a model is pre-trained on two source tasks and one target task is added. (left) In conventional multi-task learning, each task shares a common latent vector and uses a task-specific head for making predictions. During pre-training, embedding and heads for each source task are trained. Subsequently, heads for target tasks are added and trained with target data while modules trained in the pre-training stage are kept frozen. (middle) Task addition for {\gate} algorithm, which comprises embedding and task-specific modules called regression units (RU). During pre-training, embedding and RUs for source tasks are trained. Then, RUs for target tasks are added and trained with target data, while modules trained in the pre-training stage are kept frozen. (right) In {\gate}, the regression unit for task n consists of four modules: encoder, transfer ($\text{T n}$), inverse transfer ($\text{T}^{-1} \text{n}$), and head. In {\gate}, latent vectors for heads are not directly shared; instead, they are transferred to a universal locally flat (LF) space, enabling knowledge transfer through the alignment of geometrical shapes of source and target latent spaces.
}
\label{fig:fig1}
\end{figure*}

The fundamental architecture of {\gate} is precisely described in \cite{ko2024geometrically}, including the algorithm diagram, thorough mathematical structure, experiments, and an introduction to Riemannian differential geometry. The architecture of the task-added {\gate} is depicted in Figure ~\ref{fig:fig1}. As shown in the figure, a molecule is initially represented using the Simplified Molecular-Input Line-Entry system (SMILES), which serves as the input. It is then embedded into a real-number embedding vector. This embedding process is mandatory since the {\gate} algorithm requires infinitesimal perturbations around the given input vector. Consequently, the universal embedding vector must be capable of acquiring perturbation points independent of the choice of tasks.

The embedding vectors are then fed into the task-specific encoders to generate task-specific latent vectors, $z_n$. For this process, we utilize Directed Message Passing Neural Network (DMPNN) \cite{dmpnn} and conventional Multi-Layer Perceptron (MLP) layers. These latent vectors serve two main purposes: one is to compute prediction values for tasks, and the other is to align the latent spaces. 

To ensure model's accuracy, we first introduce a simple regression loss. After the latent vectors pass through the task-specific head network, the final predicted value should closely match to the given label. We use a simple Mean Squared Error (MSE) loss for this purpose.
\begin{gather}
l_{\mathrm{reg}} = \frac{1}{K}\sum_{i}^{K}(y_i - \hat{y}_i)^2
\end{gather}
Here, $K$, $y_i$, and $\hat{y}_i$ represent the number of target tasks to add, the target label, and the predicted value, respectively. In the multi-task extended \textit{\gate} architecture, the regression loss is summed over the entire number of tasks, resulting in the computation complexity of the algorithm being $\mathcal{O}(N^2)$. However, as the number of tasks is now restricted to the number of target tasks, the entire process becomes significantly faster.

The other, and the most important part of the algorithm, is the alignment of latent spaces. To align the geometrical shapes of these latent spaces, one should know the specific mapping relation between one another. In the mathematical description, the mapping should be formulated by coordinate transformation induced by a Jacobian at an arbitrary point.
\begin{gather}
    z'^i \equiv \sum_j \frac{\partial z'^i}{\partial z^j} z^j
\end{gather}
Deriving an analytic form of the Jacobian from data-driven methods without assuming the underlying geometry is typically impossible. We bypass this issue by predicting the transformed vector directly using a neural network. Neural networks are generally smooth and differentiable due to the backpropagation learning scheme, allowing us to assume the latent space is also smooth and differentiable. Hence, it is plausible to assume the latent space as Riemannian. The diffeomorphism invariance of Riemannian geometry ensures that a locally flat frame can be found anywhere on a manifold, which we utilize to align latent geometries.

\begin{figure*}[!bt]
\begin{center}
\includegraphics[width=0.92\linewidth]{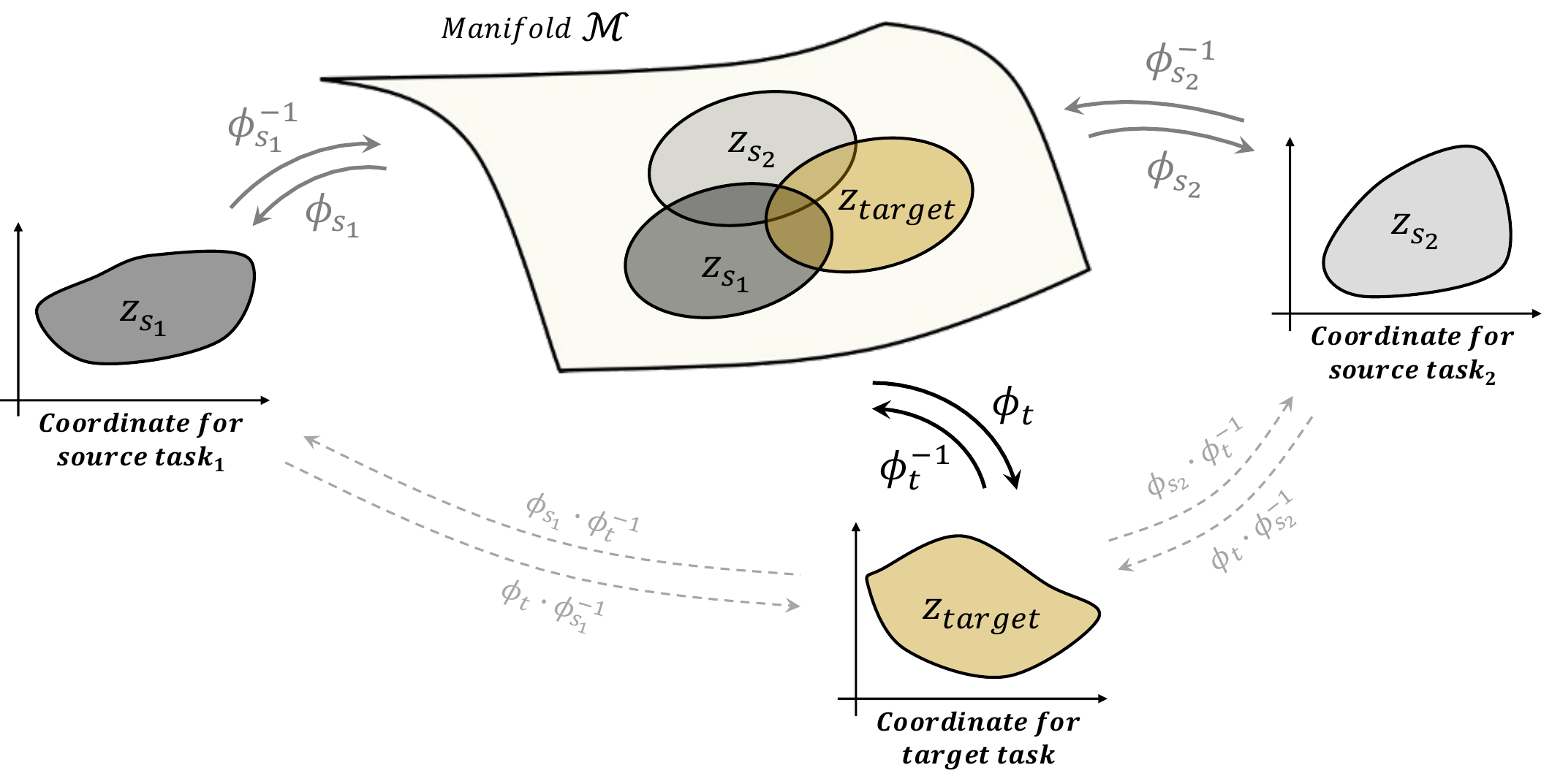}
\end{center}
\caption{Conceptual depiction of the {\gate} algorithm for task addition where the model is pre-trained on two source tasks and one target task is added. Knowledge from source tasks is transferred to a target task by aligning the geometry of the target task to the geometries of source tasks. This alignment is achieved by finding a transfer function, $\phi$, which maps an arbitrary point from a task-specific coordinate to a universal manifold, $M$. One can transform an arbitrary point in the overlapping region from one task coordinate to another by composing transfer functions. By matching the overlapping points on a manifold, one can align the inherent geometry of target data to the geometry of source data. This allows the information to flow from the source to target tasks. Grey-scaled regions are trained in the pre-training stage and frozen in the addition stage.}
\label{fig:fig2}
\end{figure*}

We set up autoencoder models to map a vector in latent space to a locally flat frame on a universal manifold and vice versa, as depicted in Figure ~\ref{fig:fig2}. Specifically, each encoder, $\phi$ maps a latent vector from a task-specific latent space to a vector in a locally flat frame, while each decoder, $\phi^{-1}$, maps a vector from the locally flat frame to the task-specific latent space. Unlike the original {\gate}, since this situation involves task addition, the models for source tasks should be pre-trained. Therefore, the mapping models can be categorized into two types: those with fixed parameters and those with learnable parameters.
Mapping from source tasks should be fixed.
\begin{gather} \label{transform_pretrained}
z'_\alpha = \mathrm{Transfer}_{\alpha\rightarrow LF}(z_\alpha)\\
\hat{z}_\alpha = \mathrm{Transfer}^{-1}_{LF \rightarrow \alpha}(z'_\alpha)
\end{gather}
And mapping from the target task should be learnable.
\begin{gather} \label{transform_learn}
z'_t = \mathrm{Transfer}_{t\rightarrow LF}(z_t)\\
\hat{z}_t = \mathrm{Transfer}^{-1}_{LF \rightarrow t}(z'_t)
\end{gather}
Here, $\alpha$ denotes the source task, and $t$ indicates the target task. The index $\alpha$ ranges from $1$ to the number of source tasks. For example, if there are $1$ to $10$ different source tasks and one new target task is added, then $\alpha$ ranges from $1$ to $10$.

Considering $\mathrm{Transfer}_{\alpha \rightarrow LF}(z_\alpha)$ as an example, where $\alpha$ is set to $4$, this indicates that the latent vector of the 4th task is mapped to a locally flat frame on a universal manifold. One can introduce a simple autoencoder loss that forces the input latent vector and reconstructed vector to be equal for the mapping networks.
\begin{equation}
l_{\mathrm{auto}} = \sum_\alpha \mathrm{MSE}(z_\alpha, \hat{z}_\alpha)
\end{equation}
Additionally, a different set of losses can be formulated to align the geometrical shapes of latent spaces. So far, we have not imposed any constraints on a mapping network that should map a latent vector to a locally flat frame. As introduced in  \cite{ko2024geometrically}, we will introduce three different kinds of constraints afterward.

One loss involves matching latent vectors on the locally flat frame that are mapped from different task-specific latent spaces. This loss aims to align the geometry point-wise by requiring latent vectors on a locally flat frame to be equal to one another. This loss is simply induced by the fact that the input of the model always starts from the same molecule regardless of the task choice. Hence, while the task-specific latent vectors may differ, if the mapping encoder correctly maps the vectors to the locally flat frame on the universal manifold, the mapped vectors from different tasks should be equal on the locally flat frame. We call this loss the consistency loss.
\begin{equation}
    l_{cons} = \sum_\alpha \mathrm{MSE}(z'_\alpha, z'_t)
\end{equation}
As mentioned earlier, $z'_\alpha$ and $z'_t$ represent vectors on the locally flat frame from the source and target latent spaces, respectively. One can introduce another loss function, namely the mapping loss, to enhance the bonding between latent spaces. This loss requires the predicted value for a given task using a standard downstream route to be equal to the predicted value for the same task using a detoured route.
\begin{gather}
    z'_\alpha = \mathrm{Transfer}_{\alpha\rightarrow LF}(z_\alpha)\\
    \hat{z}_{\alpha \rightarrow t} = \mathrm{Transfer}^{-1}_{\phantom{-1}LF \rightarrow t}(z'_\alpha)
\end{gather}
Here, $\hat{z}_{\alpha \rightarrow t}$ indicates a latent vector of the target task mapped from the vector in the source task latent space by the source mapping and target inverse-mapping networks. Although $\hat{z}_{\alpha \rightarrow t}$ originates from the source task latent space, the source mapping network should map the vector to a locally flat frame, and the target inverse-mapping network should theoretically map the vector to the same latent vector in the target latent space. Therefore, if $z_t$ and $\hat{z}_{\alpha \rightarrow t}$ go through the same head network, the predicted value should be the same. This loss is called the mapping loss.
\begin{equation}
    l_{map} = \sum_\alpha \mathrm{MSE}(y_t, \hat{y}_{\alpha \rightarrow t})
\end{equation}
$y_t$ represents a predicted value from a standard downstream route, and $\hat{y}_{\alpha \rightarrow t}$ indicates a predicted value from a detoured route.

Despite these losses, geometrical alignment remains insufficient as they only align geometry at specific points. In general, neural networks often enjoy a vast number of degrees of freedom, providing enough flexibility to significantly distort the geometrical shape. Hence, even though the geometry is matched at given points, the surrounding shape around those points may still not be aligned.

To impose stronger constraints, one should consider not only a specific point but also the geometrical shape around a given input point. Typically, this is done by assuming an analytic form of the metric and computing Riemannian properties such as curvatures. However, finding the analytic form of the metric is nearly impossible. Therefore, we will impose geometrical constraints without knowing the analytic form of the metric.

The key notion is the distance between points on a curved space. While a distance can be easily defined in Euclidean space, it is not as straightforward in curved space. One should solve the geodesic (freely falling motion) equation to find the geodesic path between points and integrate the infinitesimal distances along the path with metric weighting.
\begin{equation} \label{dist_curv}
    S^2 = \int_l \sum_\mu \sum_\nu g_{\mu\nu}dx^\mu dx^\nu
\end{equation}
However, we now know that one can always find a locally flat frame on a Riemannian manifold. By utilizing this fact, if one is trying to compute the distance between a point on a locally flat frame and a its infinitesimal perturbations, the distance is simplified into a mere Euclidean distance.
\begin{equation}
\begin{array}{ll}
    S^2 &= \int_l \sum_\mu \sum_\nu g_{\mu\nu}dx^\mu dx^\nu\\
    &= \int_l \sum_\mu \sum_\nu \eta_{\mu\nu}dx^\mu dx^\nu\\
    &= \int_a^b dx^2
\end{array}
\end{equation}
Here, $a$ represents a given latent vector, while $b$ denotes a perturbed vector around the latent vector $a$. As we assume the perturbation is infinitesimal, the distance between a given point and its perturbations can be reduced to the following form:
\begin{equation}
    S  =  |b - a|
\end{equation}
This distance should also be the same for both source and target tasks after mapping into a locally flat frame. The loss that ensures this equality is called the distance loss.
\begin{equation}
    l_{dis} = \frac{1}{M}\sum_\alpha C_\alpha\sum_{i}^M \mathrm{MSE}(s^{i}_\alpha, s^{i}_t)
\end{equation}
Here, $M$ is the number of perturbation points around the input vector, $C_\alpha$ is the weight of distance losses according to the source tasks, and $s^i_s$ is the displacement between given data and their perturbations. Specific descriptions of loss terms are as follows.
\begin{gather}
    s^{i}_\alpha \equiv |(z'_\alpha) - (z'^i_\alpha)| \qquad s^{i}_t \equiv |(z'_t) - (z'^i_t)| \\
    z'^i_\alpha = \mathrm{Transfer}_{\alpha\rightarrow LF}(\mathrm{Encoder}_\alpha(x^i))\\
    z'^i_t = \mathrm{Transfer}_{t\rightarrow LF}(\mathrm{Encoder}_t(x^i))   
\end{gather}
The index $i$ denotes $i$th perturbation point.

The distance loss is the key term in this architecture. This loss restricts the local geometrical shape around input vector points, significantly reducing the vast number of degrees of freedom of the model. As mentioned earlier, the latent space should be smooth and differentiable, limiting the freedom to retain the shape of local geometry around input vectors and preventing drastic deformations\footnote{For further information, check experiments and ablation section in \cite{ko2024geometrically}}.

Gathering all the introduced loss terms gives rise to a total loss function for the architecture.
\begin{equation}
    l_{tot} = l_{reg} + \alpha l_{auto} + \beta l_{cons} + \gamma l_{map} + \delta l_{dis}
\end{equation}
The total loss contains numerous hyperparameters that need to be tuned. These parameters affect the prediction performance of the model. Hence, some of the hyperparameters should be tuned with care to achieve superior performance. For instance, parameters $\gamma$, $\delta$, and $C_\alpha$ are sensitive to performance, while others are not. Therefore, we often leave the other parameters as $1$ and focus on finding the best set of parameters for $\gamma$, $\delta$, and $C_\alpha$. If the relationship between tasks is well known by domain knowledge, it is a very good strategy to begin with.




\section{Experiments}
\subsection{Experimental Setup}
\subsubsection{Datasets}
To evaluate our algorithm, we used 20 datasets, each corresponding to a different molecular property, sourced from three open databases: PubChem \cite{10.1093/nar/gkac956}, Ochem \cite{sushko2011online}, and CCCB \cite{cccb}. Each property was standardized by subtracting the mean and dividing by the standard deviation. Detailed information about each dataset is provided in Appendix \ref{appendix:datasets}.
We split each dataset into training and test sets using an 80:20 ratio. For evaluating model performance for extrapolation, we split the dataset based on the scaffold \cite{bemis1996properties}. The training set was further uniformly split into four folds for cross-validation. 

\subsubsection{Models}
We employed ten tasks as source tasks for the supervised pre-training of {\gate} and MTL. The remaining ten tasks were used as target tasks for task addition, individually. For each target task, we added the regression unit for the target task to the pre-trained model, resulting in ten different models for {\gate}, each for a target task.

To benchmark the performance of {\gate}, we also trained MTL models in task addition setup. In the pre-training stage, a pre-trained MTL model with a shared encoder and task-specific heads for source tasks were trained. Then, for each target task, task-specific head was added to the model architecture and trained on the target task while the pre-trained parameters were frozen. We also trained ten task-added MTL models, each for a target task.

For performance comparison, we trained {\gate} and MTL on all 20 tasks from scratch (referred to as 'Vanilla20' models), serving as reference performance for task-added {\gate} and MTL. To compare computational costs, we trained 10 MTL and {\gate} models from scratch (referred to as 'Vanilla11'), trained on 11 tasks (10 source tasks + one target task). 

In summary, we trained the following three different types of models for both {\gate} and MTL:
\begin{itemize}
    \item Task-added model: Pre-trained on 10 source tasks followed by addition and training of one target task.
    \item Vanilla20: Trained on all 20 tasks from scratch.
    \item Vanilla11: Trained on 10 source tasks and one target task from scratch.
\end{itemize}

Lastly, we also trained single-task models ('SINGLE') on each target task without multi-task learning. The same architecture for encoders and heads was used across all models for fair comparison. A detailed descriptions of the model architecture and hyperparameters are provided in Appendix Table \ref{network} and \ref{hyperparameter}, respectively.


\subsection{Results}
\subsubsection{Task Addition Reduces Computational Costs}

The primary motivation for task addition is to reduce the computational complexity of {\gate}. We analyzed training times for SINGLE, MTL, and {\gate}, as shown in Table ~\ref{table:table1}. 
Task-added MTL was the fastest to train, even taking less time than SINGLE. Task-added {\gate} was 13 \% slower than SINGLE, but 9.29 times faster than vanilla11 MTL. Task-added MTL was 10.9 times faster than vanilla11 MTL, while task-added {\gate} was 39.13 times faster than vanilla11 {\gate} 
The significant improvement in the training speed of {\gate} with task addition is due to the time complexity of vanilla {\gate} being $\mathcal{O}(N^2)$, compared to MTL's $\mathcal{O}(N)$. 
In summary, task-added {\gate} is significantly faster than training vanilla MTL or {\gate} from scratch, and its training time is comparable to that of SINGLE.

\begin{table}[h!]
\caption{Training times for SINGLE, MTL, and {\gate} algorithm.}
\label{table:table1}
\vskip 0.15in
\begin{center}
\begin{small}
\begin{sc}
\begin{tabular}{lr}
\toprule
Model              & \multicolumn{1}{l}{time per epoch{[}s{]}}  \\
\midrule
SINGLE             & 5.40                                       \\
Task-added MTL     & 5.12                                       \\
Task-added {\gate} & 6.09                                       \\
Vanilla11 MTL      & 56.00                                      \\
Vanilla11 {\gate}  & 238.32                                     \\
\bottomrule
\end{tabular}
\end{sc}
\end{small}
\end{center}
\vskip -0.1in
\end{table}

\subsubsection{Task-Added {\gate} Enables Knowledge Transfer}
\begin{figure}[!hbt]
\begin{center}
\centerline{\includegraphics[width=0.8\columnwidth]{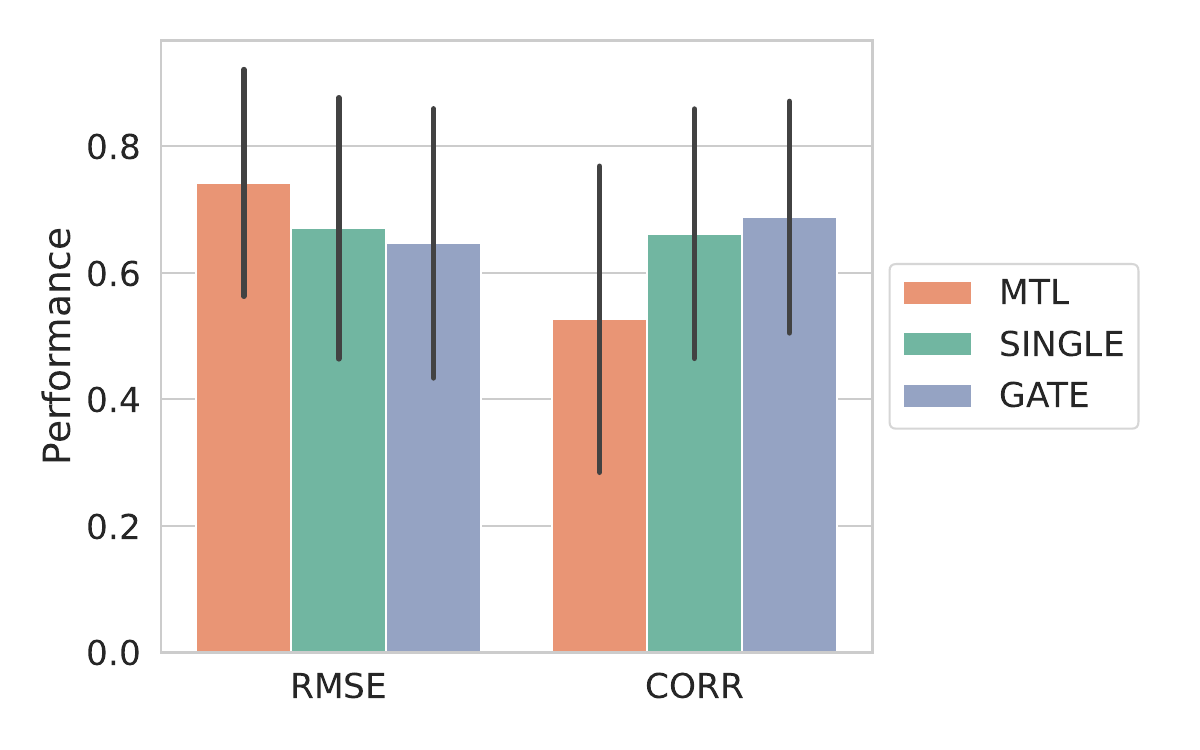}}
\caption{The performance of task-added MTL, SINGLE, and task-added {\gate} algorithm. Average RMSE and Pearson correlation values are displayed across 10 target tasks, with error bars indicating standard deviation. Detailed performance values can be found in Appendix Table \ref{performance}.}
\label{fig:fig3}
\end{center}
\vskip -0.2in
\end{figure}

The other goal of task addition is to maintain model performance compared to training the entire model from scratch. The performance of SINGLE, task-added MTL, and task-added {\gate} is illustrated in Figure ~\ref{fig:fig3}. The average Root Mean Squared Error (RMSE) of task-added MTL, SINGLE, and task-added {\gate} across 10 target tasks are 0.742, 0.670, and 0.647, respectively. Similarly, the average Pearson correlation of task-added MTL, SINGLE, and task-added {\gate} across 10 target tasks are 0.527, 0.662, and 0.688, respectively. Detailed numerical results can be found in Appendix Table \ref{performance}. These metrics demonstrate that {\gate} outperforms SINGLE, indicating that knowledge transfer through task addition enhances performance.

In contrast, MTL underperforms compared to SINGLE, suggesting that not all MTL architectures are suitable for task addition. This observation aligns with findings reported in previous studies \cite{10.1038, Deng2023}. Overall, these results indicate that {\gate} generalizes well in an extrapolation setting by using task-specific latent vector for each task, minimizing interference that might be caused by strongly shared latent vector in MTL. 

\begin{figure}[!ht]
\begin{center}
\centerline{\includegraphics[width=0.8\columnwidth]{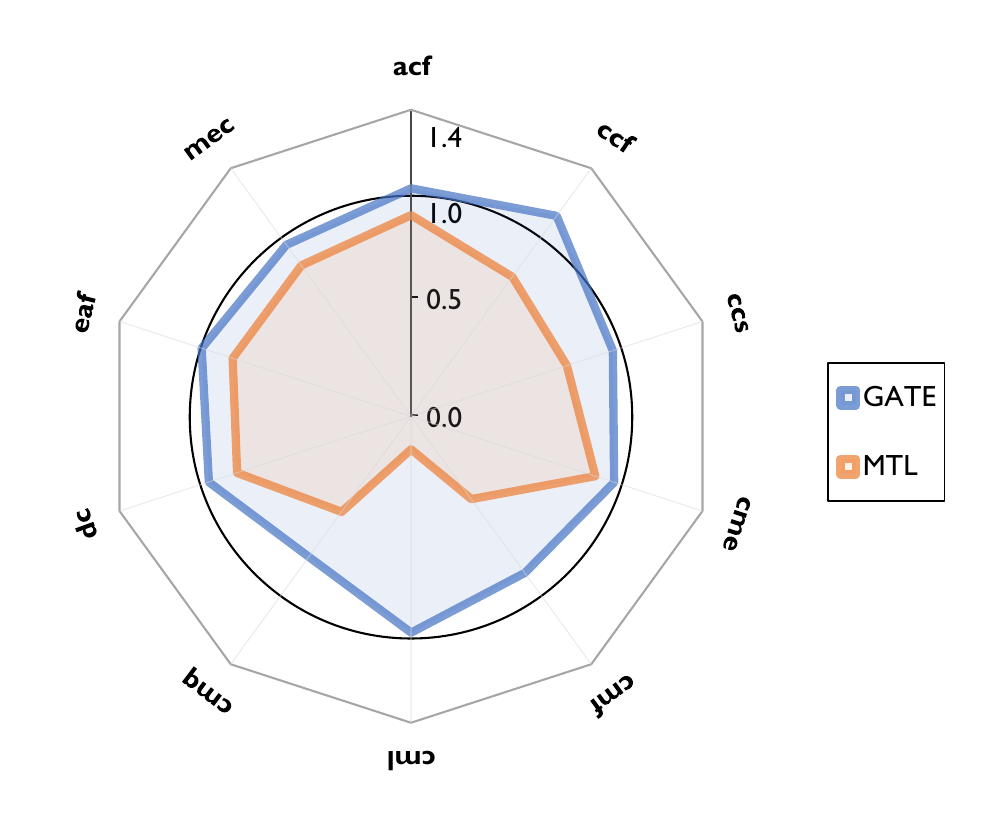}}
\caption{Correlation recovery rate of MTL and {\gate} algorithm. The correlation of task-added {\gate} and MTL is divided by the correlation of vanilla20 {\gate} and MTL, respectively. The complete names of the abbreviated tasks are listed in Appendix Table \ref{datasets}.}
\label{fig:fig4}
\end{center}
\vskip -0.2in
\end{figure}

This effect is further highlighted by comparing the correlation recovery rates of {\gate} and MTL in Figure ~\ref{fig:fig4}. On average, task-added {\gate} recovered 98.3 \% of the Pearson correlation of vanilla20 {\gate}, while task-added MTL achieved 70.4 \% of the Pearson correlation of vanilla20 MTL. {\gate} shows higher correlation recovery rates than MTL for all ten tasks. In the worst-case scenario, task-added {\gate} achieves a recovery rate of 79.3 \% for CMQ, whereas task-added MTL only recovers 15.1 \% for CML. This highlights that task-added {\gate} shows robust performance regardless of the target task, whereas the performance of task-added MTL heavily depends on the nature of the target task.
In summary, these results demonstrate that {\gate} effectively transfers knowledge from source tasks, while MTL shows limited performance in task addition setup.

\subsubsection{Dependence on Source Tasks}

\begin{figure}[!htb]
\begin{center}
\centerline{\includegraphics[width=0.8\columnwidth]{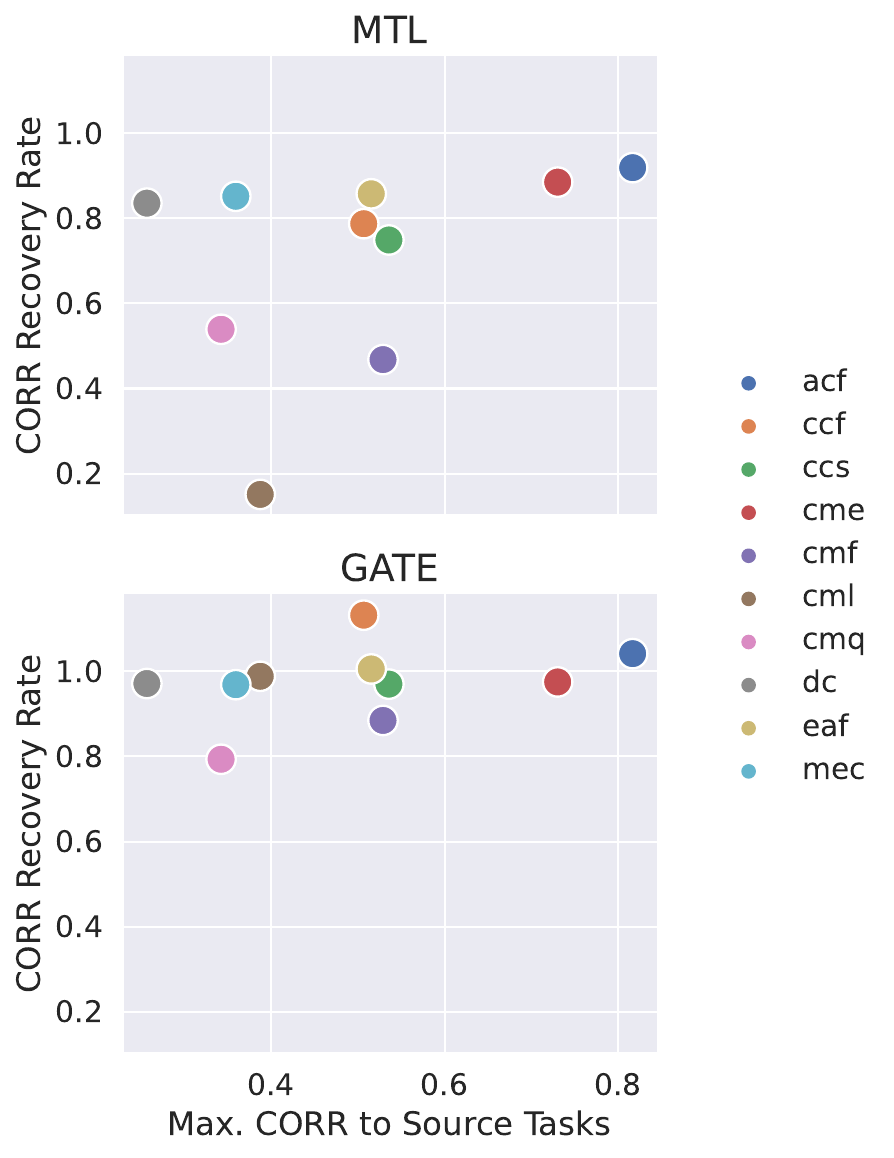}}
\caption{The relationship between correlation to source tasks and task addition performance. The performance of task-added MTL decreases as the maximum absolute correlation to the source tasks decreases, whereas the performance of task-added {\gate} is less dependent on the correlation to the source tasks. The complete names of the abbreviated tasks are provided in Appendix Table \ref{datasets}.}
\label{fig:fig5}
\end{center}
\vskip -0.2in
\end{figure}

To analyze the factors that affect the performance of task addition, we examined the correlation between source tasks and target tasks.
Specifically, for each source-target task pair, we selected molecules that have labels for both source and target tasks. If a source and target task share at least ten molecules in common, we calculated the absolute Pearson correlation between their labels to measure the degree of relatedness between source and target tasks. Figure \ref{fig:fig5} reports the maximum absolute correlation of source tasks for each target task.

For MTL, the correlation recovery rate tends to decrease as the maximum correlation between the target task and the source tasks decreases. This indicates that if no source task is closely related to the target task, the performance of multi-task learning (MTL) can decrease. On the other hand, the correlation recovery rate of task-added {\gate} is less dependent on the correlation with source tasks. This explains why task-added MTL performs worst for CML, which has a maximum related source task correlation of 0.387. Conversely, the correlation recovery rate of {\gate} for CML is 0.988, achieving a 7.3 \% increase compared to SINGLE. These results demonstrate the ability of task-added {\gate} to effectively extract mutual information even in challenging situations.

Our experimental results demonstrate that task-added {\gate} maintains the performance of vanilla {\gate} trained from scratch and shows robust performance across various target tasks. which is not achievable in conventional MTL architecture.
Additionally, task-added {\gate} achieves training times comparable to SINGLE and significantly faster than training vanilla {\gate} and MTL from scratch. 
Collectively, these findings suggest that task addition is an effective approach for reducing computational costs without sacrificing performance for {\gate}.

\section{Discussion}
In this study, we proposed and evaluated a two-stage, multi-task learning approach called task addition for the {\gate} algorithm. Task addition aims to enhance performance on target tasks with limited data while minimizing computational complexity. This is achieved through leveraging supervised multi-task pre-training on a large dataset, followed by the addition of task-specific modules. Task-added {\gate} demonstrated significant performance improvements over SINGLE and task-added MTL, with more efficient training times compared to vanilla MTL and {\gate} trained from scratch.

The {\gate}'s unique approach to knowledge transfer via geometrical alignment enhances the model's performance. Unlike conventional multi-task learning (MTL), which shares latent vectors directly, {\gate} transfers knowledge through a universal locally flat (LF) space, aligning the geometrical shapes of latent spaces across tasks. This method mitigates negative transfer effects and ensures efficient information flow from source to target tasks.

In the context of molecular property prediction, where data scarcity and high dimensionality are significant challenges, the {\gate} algorithm offers a robust solution. Our findings indicate that task-added {\gate} can maintain the performance of its vanilla counterpart while achieving training times comparable to SINGLE and task-added MTL models. This suggests that {\gate} is particularly well-suited for the efficient screening of novel compounds with desired properties by adding related tasks to the model.

While our results are promising, there are areas for further investigation. We used supervised multi-task learning as a pre-training strategy. Recently, self-supervised learning has been widely adopted as a pre-training strategy, and recent studies have showed that self-supervised pre-training followed by multi-task fine-tuning can improve performance \cite{xu2024towards}. Adopting such strategies to {\gate} may further enhance the model's performance.

In summary, the proposed task addition approach for {\gate} offers a powerful tool for improving performance on target tasks with limited data in molecular property prediction. By effectively balancing computational efficiency and task performance, the {\gate} algorithm stands out as a promising candidate for addressing the challenges of data scarcity in this domain.


\bibliographystyle{unsrt}
\bibliography{main}

\appendix
\newpage

\section{Detailed Explanation of Datasets}
\label{appendix:datasets}

We evaluated task addition using 20 molecular property datasets. During preprocessing, we excluded data with incorrect units, typographical errors, and measurements taken under extreme conditions. Before training, we standardized all datasets by their mean and standard deviation. Below, we provide the physical meaning of each molecular property.

\begin{itemize}
    \item {\bf Acentric Factor (ACF)} : A measure of the non-sphericity of molecules, which quantifies how much the behavior of a fluid deviates from that of a spherical molecule
    \item {\bf Absorbance Maximum Wavelength (AW)} : The wavelength at which a substance absorbs the maximum amount of light
    \item {\bf Boiling Point (BP)} : The temperature at which a compound changes state from liquid to gas at a given atmospheric pressure.
    \item {\bf Critical Compressibility Factor (CCF)} : A dimensionless number that describes the behavior of a substance at its critical point.
    \item {\bf Collision Cross Section (CCS)} : A measure of the probability of interaction between particles, representing the effective area that one particle presents to another for a collision to occur.
    \item {\bf Chromophore Emission Max (CME)} : The wavelength at which a chromophore emits the maximum amount of light upon excitation.
    \item {\bf Chromophore Emission FWHM (CMF)} : The width of the emission spectrum at half of its maximum intensity, indicating the range of wavelengths emitted by the chromophore.
    \item {\bf Chromophore Life Time (CML)} : The average time a chromophore remains in an excited state before returning to its ground state.
    \item {\bf Chromophore Quantum Yield (CMQ)} : The efficiency of photon emission by a chromophore, defined as the ratio of the number of photons emitted to the number of photons absorbed.
    \item {\bf Decomposition (DC)} : The process by which a chemical compound breaks down into simpler substances or elements.
    \item {\bf Dipole Moment (DM)} : A measure of the separation of positive and negative charges in a molecule, indicating the polarity of the molecule.
    \item {\bf Electron Affinity (EAF)} : The amount of energy released when an electron is added to a neutral atom or molecule in the gas phase to form a negative ion.
    \item {\bf Flash Point (FP)} : The lowest temperature at which a liquid can form an ignitable mixture in air near its surface, indicating its flammability.
    \item {\bf HOMO (HM)} : The highest energy molecular orbital that is occupied by electrons in a molecule under normal conditions.
    \item {\bf LUMO (LM)} : The lowest energy molecular orbital that is unoccupied by electrons, which can accept electrons during a chemical reaction.
    \item {\bf Log P (LP)} : The logarithm of the partition coefficient between octanol and water, indicating the hydrophobicity or lipophilicity of a compound.
    \item {\bf Molar Extinction Coefficient (MEC)} : A measure of how strongly a chemical species absorbs light at a given wavelength per molar concentration.
    \item {\bf Melting Point (MP)} : The temperature at which a solid turns into a liquid at atmospheric pressure.
    \item {\bf pKa (PKA)} : The negative logarithm of the acid dissociation constant, indicating the strength of an acid in solution.
    \item {\bf Refractive Index (RI)} : A measure of how much light is bent, or refracted, when entering a material from another medium.

\end{itemize}

\begin{table*}[!bt]
\caption{Dataset statistics}
\label{datasets}
\vskip 0.15in
\begin{center}
\begin{small}
\begin{sc}
\begin{tabular}{llllrrr}
    \toprule
    \textbf{Name} & \textbf{Acronym} & \textbf{Type} & \textbf{Count} & \textbf{Mean} & \textbf{STD} \\
    \midrule
    Acentric Factor                 & ACF & Addition   & 1850  & -0.33  & 0.23   \\
    Absorbance Maximum Wavelength   & AW  & Pretraining & 11896 & 440.39 & 67.80  \\
    Boiling Point                   & BP  & Pretraining & 8044  & 183.65 & 96.24  \\
    Critical Compressibility Factor & CCF & Addition   & 1357  & 0.25   & 0.03   \\
    Collision Cross Section         & CCS & Addition   & 4006  & 205.06 & 57.84  \\
    Chromophore Emission Max        & CME & Addition   & 6407  & 504.95 & 100.38 \\
    Chromophore Emission FWHM       & CMF & Addition   & 2862  & 72.14  & 26.04  \\
    Chromophore Life Time           & CML & Addition   & 2738  & 0.49   & 0.64   \\
    Chromophore Quantum Yeild       & CMQ & Addition   & 5609  & 0.35   & 0.29   \\
    Decomposition                   & DC  & Addition   & 12998 & 210.86 & 58.31  \\
    Dipole Moment                   & DM  & Pretraining & 11224 & 0.30   & 0.53   \\
    Electron Affinity               & EAF & Addition   & 198   & 15.24  & 28.36  \\
    Flash Point                     & FP  & Pretraining & 9409  & 114.21 & 82.87  \\
    HOMO                            & HM  & Pretraining & 97262 & -5.66  & 0.64   \\
    LUMO                            & LM  & Pretraining & 97262 & -1.63  & 0.89   \\
    Log P                           & LP  & Pretraining & 31264 & 10.33  & 9.79   \\
    Molar Extinction Coefficient    & MEC & Addition   & 16324 & 7.67   & 0.64   \\
    Melting Point                   & MP  & Pretraining & 22901 & 376.53 & 92.31  \\
    pKa                             & PKA & Pretraining & 9514  & 6.62   & 3.11   \\
    Refractive Index                & RI  & Pretraining & 11143 & 1.49   & 0.10 \\  
    \bottomrule
\end{tabular}
\end{sc}
\end{small}
\end{center}
\vskip -0.1in
\end{table*}


\section{Architecture and Hyperparameters}
Training process for {\gate} is described in  \cite{ko2024geometrically}. Our model is composed of five different types of modules, whose parameter sizes are listed in Table \ref{network}.

\begin{table*}[!hb]
\caption{Network parameters}
\label{network}
\begin{center}
\begin{small}
\begin{sc}
    \begin{tabular}{ccccc}
    \toprule
    \textbf{network} & \textbf{layer} & \textbf{input, output size} & \textbf{hidden size} & \textbf{dropout} \\ 
    \midrule
    backbone         & DMPNN     & [134,149], 100 & 200& 0    \\
    bottleneck       & MLP layer & 100, 50        & 50 & 0    \\
    transfer         & MLP layer & 50, 50         & 50 & 0.2  \\
    inverse transfer & MLP layer & 50, 50         & 50 & 0.2  \\
    head             & MLP layer & 50, 1          & -  & 0.2  \\
    \bottomrule
    \end{tabular}
\end{sc}
\end{small}
\end{center}
\vskip 0.2in
\end{table*}


The hyperparameters for training are described in Table \ref{hyperparameter}.
\begin{table}[htbp]
\caption{Hyperparameters}
\label{hyperparameter}
\begin{center}
\begin{small}
\begin{sc}
    \begin{tabular}{cc}
    \toprule
    \textbf{Hyperparameter} & \textbf{Value} \\
    \midrule
        learning rate                  & 0.0005  \\
        optimizer                      & AdamW  \\
        batch size                     & 512  \\
        epoch                          & 1000  \\
        \# of perturbation             & 5  \\
       $\alpha, \beta, \gamma, \delta$ & 1, 1, 1, 1  \\
    \bottomrule
    \end{tabular}
\end{sc}
\end{small}
\end{center}
\end{table}


\newpage
\section{Experimental Results}
Here, we describe the explicit test results in this section. The numerical results for Figure \ref{fig:fig3} is reported in Table \ref{performance}.

\begin{table}[!htb]
\caption{The performance of SINGLE, task-added MTL, and task-added \gate algorithm}
\label{performance}
\begin{center}
\begin{small}
\begin{sc}
\begin{tabular}{lllll}
    \toprule
           & \multicolumn{2}{l}{RMSE}          & \multicolumn{2}{l}{Correlation}   \\
           & \multicolumn{1}{l}{Mean}  & STD   & \multicolumn{1}{l}{Mean}  & STD   \\
    \midrule
    Task-Added MTL    & \multicolumn{1}{l}{0.742} & 0.179 & \multicolumn{1}{l}{0.527} & 0.242 \\
    SINGLE               & \multicolumn{1}{l}{0.670} & 0.206 & \multicolumn{1}{l}{0.662} & 0.197 \\
    Task-Added \gate  & \multicolumn{1}{l}{0.647} & 0.213 & \multicolumn{1}{l}{0.688} & 0.183 \\
    \bottomrule
\end{tabular}
\end{sc}
\end{small}
\end{center}
\vskip 0.2in
\end{table}

The correlation recovery rate depicted in Figure \ref{fig:fig4} are listed in Table \ref{table:corr_recovery}.

\begin{table}[!htb]
\caption{The correlation recovery rate of MTL, and {{\gate}} algorithm for each target task.}
\label{table:corr_recovery}
\begin{center}
\begin{small}
\begin{sc}
\begin{tabular}{lll}
\toprule
Task & MTL   & GATE  \\
\midrule
acf  & 0.919 & 1.041 \\
ccf  & 0.787 & 1.132 \\
ccs  & 0.749 & 0.970 \\
cme  & 0.885 & 0.975 \\
cmf  & 0.468 & 0.884 \\
cml  & 0.151 & 0.988 \\
cmq  & 0.539 & 0.793 \\
dc   & 0.835 & 0.971 \\
eaf  & 0.857 & 1.005 \\
mec  & 0.852 & 0.968 \\
\bottomrule
\end{tabular}
\end{sc}
\end{small}
\end{center}
\vskip -0.1in
\end{table}

\end{document}